\documentclass[runningheads]{llncs}
\usepackage{iftex}
\ifPDFTeX
  \usepackage[utf8]{inputenc} 
  \usepackage[T1]{fontenc}
  \usepackage{lmodern}
  \usepackage{newtxmath,newtxtext}
\else
  \usepackage{fontspec}
  \usepackage{microtype}
\fi
\usepackage[english]{babel}
\usepackage{verbatim}
\usepackage{tabularx}
\usepackage{booktabs}
\usepackage{graphicx}
\usepackage[hyphens]{url}
\usepackage[normalem]{ulem}
\usepackage[hidelinks]{hyperref}
\usepackage{linguex}

\usepackage[group-separator={,},group-minimum-digits=4]{siunitx}
\usepackage{multirow}
\newcommand{\exref}[1]{\\[3pt]\mbox{}\hfill #1}
\begin{document}
\title{From \textit{Plenipotentiary} to \textit{Puddingless}: Users and Uses of New Words in Early English Letters\texorpdfstring{\thanks{Licence: Creative Commons Attribution 4.0 International (CC BY 4.0)}}{}}
\titlerunning{Users and Uses of New Words in Early English Letters}
%
\author{Tanja Säily\orcidID{0000-0003-4407-8929} \and
Eetu Mäkelä\orcidID{0000-0002-8366-8414} \and \\
Mika Hämäläinen\orcidID{0000-0001-9315-1278}}
\authorrunning{T. Säily et al.}
%
\institute{University of Helsinki, Finland
\email{\{tanja.saily,eetu.makela,mika.hamalainen\}@helsinki.fi}}
\maketitle              
\begin{abstract}
We study neologism use in two samples of early English correspondence, from 1640--1660 and 1760--1780. Of especial interest are the early adopters of new vocabulary, the social groups they represent, and the types and functions of their neologisms. We describe our computer-assisted approach and note the difficulties associated with massive variation in the corpus. Our findings include that while male letter-writers tend to use neologisms more frequently than women, the eighteenth century seems to have provided more opportunities for women and the lower ranks to participate in neologism use as well. In both samples, neologisms most frequently occur in letters written between close friends, which could be due to this less stable relationship triggering more creative language use. In the seventeenth-century sample, we observe the influence of the English Civil War, while the eighteenth-century sample appears to reflect the changing functions of letter-writing, as correspondence is increasingly being used as a tool for building and maintaining social relationships in addition to exchanging information.

\end{abstract}

\section{Introduction}

In this paper, we report on our work to study novel vocabulary, or neologisms, in a corpus of historical letters, namely the \textit{Corpora of Early English Correspondence} (CEEC)~\cite{ceec}. The CEEC was specifically designed for the purposes of historical sociolinguistics, and as such it 1) aims at social representativeness in terms of e.g. gender and social rank and 2) records this social metadata for each letter included. Previous research into the history of English lexis has focused on published texts, which were mostly written by highly educated men. By analysing personal correspondence, a genre that was accessible to anyone who was literate, we are able to consider the language use of women and the lower social ranks as well. In this study, we are especially interested in the early adopters of new vocabulary, the social groups they represent, and the types and functions of their neologisms.

However, the varied nature of the corpus also poses problems for neologism identification. The CEEC compilers wanted to study authentic language use, and thus the corpus is compiled from editions that retain original spelling. English spelling did not develop a prescribed standard until the eighteenth century, and there is a great deal of spelling variation in historical English texts -- and in the CEEC. In total, the CEEC contains about \num{150000} distinct word forms. By excluding those tagged as foreign, proper nouns or abbreviations, about \num{125000} remain. Our initial idea was to take this vocabulary, and match it against the Oxford English Dictionary (OED)~\cite{oed} and the Middle English Dictionary (MED)~\cite{med} to discover which words appeared in the CEEC earlier than mentioned in the dictionaries. However, even though the OED and MED contain some \num{375000} and \num{260000} spelling variants for their \num{280000} and \num{60000} words, only some \num{36000} word forms from the CEEC could be directly mapped to the dictionaries, while \num{88000}, or 70\%, could not. Further, the proportions of words that could be mapped differed widely between social ranks, with words used by the lower ranks having a significantly lower chance of finding a match in the dictionaries. This in turn made straightforward statistical comparisons between social ranks impossible.\footnote{For more on the problematic nature of such non-standard data for analysis, see~\cite{makela2020wrangling}.} To try to counteract these problems, we turned to automatic normalization approaches, which we hoped would allow us to map a significantly larger portion of the words against the two dictionaries.
 
\section{Related Work}

There is a lot of recent NLP research conducted on historical data, especially in meaning change over long periods of time. On the other hand, there are also recent studies on neologisms in present-day data. In this section, we describe some of the recent related work.

In linguistics, previous research on historical neologisms has typically gathered its data from dictionaries, which are biased towards well-known authors, while corpus-based studies have chiefly concentrated on individual derivational affixes~\cite{Nevalainen1999,Palmer2015}. Present-day studies have used more automated methods to discover a potentially wider range of neologisms in massive corpora compiled from e.g. newspapers or tweets~\cite{Renouf2007,Grieve2017}, but they have tended to ignore social categories beyond regional variation.

A recent study shows a method for normalizing neologisms in social media texts~\cite{zalmout-etal-2019-unsupervised}. Their method tries to first detect a neologism and then normalize it into standard language. Interestingly, they apply the idea of normalization in order to remove neologisms, where as we use normalization to find neologisms \cite{hamalainen-etal-2019-revisiting}.

Another neologism identification study focuses on multi-word neologisms, namely adjective-noun pairs \cite{mccrae-2019-identification}. Their approach uses BERT \cite{devlin2019bert} and ELMo \cite{peters2018deep} in detecting whether a given adjective-noun pair is a neologism. Our work focuses on individual words that are neologisms as opposed to multi-word expressions.

Spanish neologisms have been studied in a corpus of conversations of language learners \cite{wein-2020-classification}. They study neologisms in a neologism-annotated corpus examining the relationships between neologisms, loanwords, and errors. Loanwords and errors are typically produced by language learners and true neologisms by native speakers. Unlike our corpus, their corpus is annotated for neologisms.

Lexical semantic change has been studied with word embeddings \cite{dubossarsky2019time}. The authors suggest that word embeddings should not be aligned when used to study semantic change, as such an alignment introduces noise. Their approach introduces less noise which ultimately leads to more accurate results. Despite this, the use of machine learning approaches to study the phenomenon is not free of problems \cite{hengchen2021challenges}.

Normalization is a very common practice embraced in the NLP research when dealing with historical or otherwise non-standard language \cite{domingo2020modernizing,chakravarthi2020survey,partanen2019dialect,duong2020unsupervised,zarnoufi2020manorm}. The benefit of normalization is that it makes non-standard orthography standard and thus enables the use of NLP tools and resources designed for modern normative data.

\section{Neologism Pipeline}

The details of our automated mapping procedure are described in \cite{hamalainen2018normalizing}. In short, the best-performing approach we could come up with was based on neural machine translation, with a post-filtering step that accepted only lemmas appearing in the OED~\cite{hamalainen-etal-2019-revisiting}. Access to a local version of the OED was kindly provided by Oxford University Press. Overall, this approach was able to recover accurate lemmas for \num{61}\% of previously unmatched words. 

However, a broader evaluation of the whole pipeline cast a dark shadow over the overall usability of the method. Out of all words in a sample of 17th-century words, \num{75}\% were matched to a lemma. However, for a full \num{30}\% out of those, that lemma turned out to be a wrong one! Finally, even out of the words that were matched to the right lemma, \num{17}\% were matched to the wrong part of speech, which, when compared to the OED, could equate to a possibly wrong earliest attestation date. Actually, for this data, even if the lemmatization were perfect, about \num{20}\% of the words would still get mapped to the wrong entry without further disambiguation based on part of speech or other contextual information. On top of this, the normalization accuracy was not the same for each social category. While, interestingly, lemmatization accuracy was similar across social ranks, it differed according to gender, as well as according to the type of correspondence (between family members, between friends or between more distant acquaintances).

In light of this, while we originally endeavoured to come up with a mostly automated pipeline that could be used to study neologisms en masse and quantitatively, in the end we were forced to conclude that this would not be possible. Instead, we resigned ourselves to treat the automatically derived lemmas and OED matches as only suggestions that would be verified manually. This in turn meant a significant amount of manual work for analysing any sizable part of the corpus. We therefore adopted a time-limited sampling-based approach, where we first decided on a twenty-year time\-span of the CEEC where we had a relatively even coverage of different social categories, and then sampled neologism candidates from that period using a stratified sampling approach that tried to include as equal an amount of running words from each social category as possible.

More specifically, we decided that we wanted our corpora to be as balanced as possible on three social axes: 1) the gender of the letter writer, 2) their social rank, and 3)~register in terms of the relationship between the writer and the recipient. Based on an evaluation of the corpus balance over time along these axes, we settled on two time periods: 1640--1660 and 1760--1780. The former was also of interest because it coincided with the English Civil War. We then grouped the data from those periods according to the three criteria, and sampled letters as equally as possible from all buckets. From each sampled letter were then extracted the words that appeared for the first time in the CEEC in that letter. These then formed the candidates for neologisms without any regard yet to the dictionaries. The sizes of the samples were controlled so that each could be gone through manually in a reasonable time, leading in the end to candidate neologisms lists of \num{820} word types for the 17th century and \num{645} for the 18th century (both about 9\% of the the total sampling pool of neologism candidates, which was \num{8954} for the 17th century and \num{7131} for the 18th century).\footnote{We would like to thank Jukke Kaaronen for going through the 18th-century sample, which was then analysed further by us.}

\begin{table}[htbp]
    \caption{Number of running words in the two samples for each of the three social axes of interest.}
    \label{tab:samples}
    \centering
    \begin{tabular}{l@{\qquad}l@{\qquad}r@{\qquad}r}
    \toprule
    Category & Value & 17th c. & 18th c. \\
    \midrule
    Total & & \num{36265} & \num{47864} \\
    \midrule
    Gender & Male & \num{23459} & \num{29225} \\
    & Female & \num{12806} & \num{18639} \\
    \midrule
    Social Rank & Royalty & \num{3899} & \num{4067} \\
    & Nobility & \num{5038} & \num{6998} \\
    & Gentry & \num{11509} & \num{10924} \\
    & Clergy & \num{9659} & \num{8976} \\
    & Professionals & \num{3675} & \num{10847} \\
    & Merchants & \num{860} & \num{2496} \\
    & Other non-gentry & \num{1625} & \num{3556} \\
    \midrule
    Relationship & Nuclear family & \num{15045} & \num{12754}\\
    & Other family & \num{0} & \num{6534} \\
    & Close friends & \num{7467} & \num{14771} \\
    & Other acquaintances & \num{13753} & \num{13805} \\
    \bottomrule
    \end{tabular}
\end{table}

Given the limitations inherent in the corpus, this did not result in a completely balanced sample, as shown in Table \ref{tab:samples}, but at least its biases are known and can be taken into account in the analysis. 

We then used our FiCa tool~\cite{Saily2018b} to manually verify each lemma and OED association suggestion from our automated pipeline, as well as to fill in these for the words where our approach had not yielded candidate mappings. After receiving this corrected and amended mapping, we finally used it to source earliest attestation dates from the OED for each of the lemmas. For this study, we decided to define a neologism as a word whose earliest attestation date in the OED was at most forty years before its appearance in our sample. After a final round of manual verification, this procedure yielded us 42 novel word types in the 17th-century sample, with a total of 53 token-level appearances. For the 18th-century sample, we obtained 21 novel word types, each of which appeared only once.

For analysis, these neologism tokens were then combined with the social metadata associated with the letters they appeared in, but also etymological and semantic data as sourced from the OED. Here, we looked at each sense entry for each word entry and each semantic node in the Historical Thesaurus of the OED for each sense. The semantic nodes give us important information about the ontological meaning of each neologism candidate, as the Historical Thesaurus (HT) records hierarchical information about the meaning of each entry such as `society » communication » writing » handwriting or style of'. This makes it possible for us to inspect the neologisms on different levels of the hierarchy.

As the number of neologisms in our samples is so low, the data does not lend itself to statistical hypothesis testing. Thus, all numerical results from the following analysis should be taken as provisional.\footnote{The analysis dataset is available on Zenodo~\cite{saily_tanja_2021_4578940}.} 

\section{Seventeenth-Century Neologisms}

\subsection{Overview}

The 17th-century sample provided us with 53 neologism tokens representing 42 types, listed below. Antedatings to the OED (entries and senses, checked against OED Online in February 2021) are marked in boldface on the list. 

\begin{quote}
acrimonious, believingly, candid, candour, causally, compensate, compliance, condescension, coney ground, congregational, \textbf{covenanting} (adj.), \textbf{crawling}~(n.), dishearten, dragooner, \textbf{efficaciously}, eminently, endeared, entanglement, \textbf{helpfulness}, \textbf{hint} (v.), idolum, incendiary, \textbf{incognito}, initiatory, \textbf{joke} (n.), \textbf{landgravine}, \textbf{malignancy}, manifesto, \textbf{oversweetness}, \textbf{packet-boat}, \textbf{plenipotentiary} (n.), remind, rickets, sequestrator, servient (n.), \textbf{statement}, Swede, \textbf{tea}, variously, \textbf{vibrate} (v.), visit (n.), voluminous
\end{quote}

Unsurprisingly, nouns being the largest lexical category in general, most of the neologisms (24) are nouns as well, followed by adjectives (8), verbs (5) and adverbs (5). While half of the words have been formed within English through derivation (18), compounding (2) or conversion (1), there are also a large number of borrowings (19): thirteen from Latin and two each from French, Italian and German. The OED gives no certain etymology for two nouns, \textit{joke} and \textit{rickets}, although the former may come from Latin. Looking at the top level of the HT classification and the main semantic class of each word, 16 of the neologism types relate to `society', 15 to `the world' and 11 to `the mind'. Within these classes, the words are distributed across as many as 22 second-level categories, the most frequent ones being `society » authority' (5 neologism types) and `the mind » mental capacity' (4 types). The former may be at least partly connected to the Civil War, as in example \ref{ex:malignancy}, where \textit{malignancy} seems to be used in the sense `political disaffection'. Sir Anthony was a Royalist who was charged with misapprobation of public monies and imprisoned during the war.

\ex. \label{ex:malignancy}For what I made over to Mr.\ Kenrick's it was uppon reall considerations such as will appeare good if lawe have any being; but of that I will not dispute, considering my durance and being cloathed by some particular men
with the garment of \textbf{Malignencye} and therfore in a suffring condition.\exref{(Sir Anthony Percival to his neighbour, Henry Oxinden, 1643;\\\mbox{}\hfill OXINDE\_186)\footnote{Examples are given in their original spelling. Boldface has been added, while italics (if any) are as printed in the letter edition and probably reflect underlining in the original manuscript. OXINDE\_186 is the unique identifier of the letter in the corpus.}}

Words used in the context of the Civil War can be found in a number of different semantic classes, however. An obvious one is \textit{dragooner} `society » armed hostility » warrior' used by Charles~I himself, but there are also \textit{packet-boat} `society » travel » travel by water', \textit{sequestrator} `society » law » administration of justice' and \textit{statement} `the mind » language » statement', to name a few. Most of them do seem to be found under the top-level class of `society', which makes sense as the Civil War was chiefly a societal phenomenon.

\subsection{Sociolinguistic Variation}

In what follows, we will compare the normalized frequencies of neologism tokens across different social groups; see Table \ref{tab:c17freqs}. These are used as a starting point for a more qualitative analysis of the users and uses of neologisms.

\begin{table}[htbp]
    \caption{Normalized frequency of neologism tokens per \num{10000} words in the 17th-century sample for the three main social axes of interest, sorted by frequency.}
    \label{tab:c17freqs}
    \centering
    \begin{tabular}{l@{\qquad}l@{}r}
    \toprule
    Category & Value & Frequency / \\
    & & \num{10000} words \\
    \midrule
    Gender & Male & \num{17} \\
    & Female & \num{11} \\
    \midrule
    Social Rank & Royalty & \num{26} \\
    & Professionals & \num{24} \\
    & Nobility & \num{16} \\
    & Gentry & \num{13} \\
    & Clergy & \num{11} \\
    & Merchants & \num{0} \\
    & Other non-gentry & \num{0} \\
    \midrule
    Relationship & Close friends & \num{25} \\
    & Other acquaintances & \num{18} \\
    & Nuclear family & \num{6} \\
    & Other family & -- \\
   \bottomrule
    \end{tabular}
\end{table}

The two \textbf{social ranks} with the highest frequency of neologisms are royalty and professionals, both with c.\ 25 instances per \num{10000} words. Although Charles~I uses three neologism tokens (\textit{dragooner} once and \textit{dishearten} twice), most of the royalty's use of new words is due to Elizabeth Stuart, Queen of Bohemia, who lived abroad, basically in exile, at the time. She wrote about \textit{visits} and being \textit{incognito}, and discussed people ranging from \textit{servients} to \textit{landgravines}, \textit{plenipotentiaries} and \textit{Swedes}, the latter of whom were actively engaged in the Thirty Years' War. Amongst the nine professionals in the sample, new vocabulary is only used by two, Parliamentarian army officers John Dixwell (\textit{sequestrator}) and Thomas Harrison (\textit{believingly}, \textit{condescension}, \textit{endeared}, \textit{hint} v., \textit{variously}). Many estates owned by Royalists were sequestered during the Civil War, and war and religion were mixed in Harrison's discussion of Cromwell with Colonel John Jones \ref{ex:believingly}.

\ex. \label{ex:believingly}To agree (as is alreadie) to act in dearest love expressed to him named Protector, (or Mount Sirion as the Sidonians called Hermon, and David in the spirit followed that faithfully, \textbf{believingly}, undoubtingly, unanimously, that He would retreat in action of undertaking (and soe witnes repentance by \textbf{condisention}) and wee would as willingly repent of o\textsuperscript{r} sinfull dissentions) I shall therefore apply what I have now brought to offer, onely to that.\exref{(Thomas Harrison to John Jones, 1656; JONES\_040)}

As regards \textbf{gender}, men tend to use more neologisms than women. Interestingly, the women in the sample mostly use borrowings, whereas the majority of the men's neologism tokens were formed within English. The female early adopters were in general well educated, which could explain their use of the borrowings. They belonged to the social ranks of royalty (Elizabeth Stuart), nobility (Anne Conway), gentry (Brilliana Harley), and clergy (Anne Cary and Winefrid Thimelby, nuns). Lady Harley used the potentially Civil War related term \textit{incendiaries} in a political discussion with her son, who later became a Parliamentarian army officer; both her and Elizabeth's use of new vocabulary focuses on the semantic class `society » authority'. Lady Conway, on the other hand, discussed philosophical concepts with her friend and fellow philosopher, Henry More, as in \ref{ex:idolum}. The OED records More's use of \textit{idolum} in a publication in 1647, so both of the correspondents would have been familiar with the term.

\ex. \label{ex:idolum}Now the papyr certainly could not possible be capable of perceiving motion nor could it transmitt its motion from the obiect to the eye, for first the paper transmitts the colour of white w\textsuperscript{ch} is its own motion, and if it should transmitt the motion caused by any other obiect, then why does not everything we Looke upon yeeld the \textbf{eidolum} or representation of something else […]\exref{(Lady Anne Conway to Henry More, 1651; CONWAY\_093)}

Considering \textbf{register}, the frequency of neologisms is at its highest in letters written to close friends (examples \ref{ex:believingly} and \ref{ex:idolum} above) and, to a lesser extent, other acquaintances \ref{ex:malignancy}, whereas letters to family members do not seem to be particularly fertile ground for neologisms. This is true for both the men and the women in the sample.

Looking at \textbf{age}, it seems that older people (40--70) use neologisms more frequently than do younger people (17--39); the normalized frequencies are \num{21} and \num{10} per \num{10000} words, respectively. This observation holds even if the age groups are split at 50 years, shifting two of the outliers, Elizabeth Stuart and Thomas Harrison, to the younger side; in that case, the normalized frequency is \num{18} for the older group and \num{15} for the younger. The youngest user of neologisms in our sample is the above-mentioned philosopher, Lady Conway, at twenty. The oldest person in the sample is the seventy-year-old Sir Hamon L'Estrange, a Royalist politician, who uses five new words in a single letter to his physician, Dr Thomas Browne. Some of these are related to his ailment and potential cures found elsewhere (\textit{acrimonious}, \textit{oversweetness}, \textit{manifesto}), while others describe L'Estrange's own efforts in science as \textit{crawling} (n.) and expect \textit{candour} from Browne. His skilful use of recent vocabulary is not unexpected, as Kyle~\cite{LEstrangeSirHamon15831654politician} characterizes L'Estrange as a ``cultured and articulate man'' who had many interests and who accumulated a large library.

In terms of \textbf{regional variation}, the amount of data per category permits us to say very little, but going beyond the existing metadata categories, we make the impressionistic observation that people residing abroad, either permanently or more temporarily, frequently use new words, which are naturally often borrowed. This goes for Elizabeth and the two nuns mentioned above, but also e.g. two Royalist noblemen, Thomas Howard, Earl of Arundel and Surrey, and his son, William Howard, Viscount Stafford, who travelled on the continent during the war. In fact, they produced three of the most interesting neologisms in our sample: \textit{packet-boat} \ref{ex:packetboat}, \textit{statement} \ref{ex:statement} and \textit{tea} \ref{ex:tea}. Not only do these antedate the first attestations provided by the OED, but we have also been unable to find earlier instances in massive databases of contemporary published texts like Early English Books Online~\cite{eebo}.

\ex. \label{ex:packetboat}S. Jhon Pennington, just nowe Count Fabroni, \& President Cognewe are come unto me from Q: Mother, to entreate very earnestly, that the gentleman cominge alonge w\textsuperscript{th} this called Don Martino Dugaldi may instantly passe to Dunkerke for her M\textsuperscript{ties} especiall service w\textsuperscript{ch} depends soe much upon it as upon his retorne or any others sent before by y\textsuperscript{e} \textbf{Packette Boate} […]\exref{(Thomas Howard in Dover to Sir John Pennington, 1641;\\\mbox{}\hfill ARUNDEL\_068; OED3 first attestation 1642)}

\ex. \label{ex:statement}I have receaved onely one letter in which there is a \textbf{statement} that the ssouldiers went to Mr John Penneducks house at King berry and ransaked it totally […]\exref{(William Howard in Antwerp to Thomas Howard, 1642;\\\mbox{}\hfill ARUNDEL\_072; OED3 first attestation 1750)}

\ex. \label{ex:tea}I have scarce bought any thinge for my selfe but an Indian Brewhouse for \textbf{tee}, which hath beene very good Black Lack worke, but it is all spoyled and rased and yett I payed exceeding deare for it.\exref{(William Howard in Amsterdam to Aletheia Howard, 1643;\\\mbox{}\hfill ARUNDEL\_074; OED2 first attestation 1655)}

Examples \ref{ex:packetboat} and \ref{ex:statement} seem to be related to the beginnings of the Civil War. As noted by Säily et al.~\cite{Saily2019} regarding Viscount Stafford and example \ref{ex:tea},

\begin{quote}
He was writing from Amsterdam to his mother, who was also staying in the Low Countries. We're not really sure what an ``Indian Brewhouse'' is, but it’s clearly not an actual house, being lacquerware (which was frequently imported from the East Indies); rather, it seems to be some sort of a container for tea. It makes sense that being in the Low Countries, William and his mother would have known the Dutch word for tea. Furthermore, the word isn't marked in any way in the text (the boldface was added by us), which indicates that it was quite a normal word for them -- often, the new words we encounter in letters have been underlined or explained with another word as a sign of their novelty. It seems, then, that the upper social ranks of the time, at least those who travelled on the continent, could have been quite familiar with tea-drinking, so the early history of the word in English is not just about technical discussions of the plant but about everyday usage as well. 
\end{quote}

\section{Eighteenth-Century Neologisms}

\subsection{Overview}

Even though the 18th-century sample was larger than the 17th-century one, it only yielded 21 neologism types, listed below. Each of the types only occurred once in the sample. Antedatings to the OED are marked in boldface on the list, and words whose exact sense or part of speech could not be found in the OED are underlined. For the latter, we came up with suitable metadata based on neighbouring entries or senses in the OED.

\begin{quote}
anecdote-monger, blacky, \uline{cream-can}, dénouement, \uline{floreat} (n.), foundling-house, funny, \textbf{hookah}, \textbf{inspectress}, \textbf{interference}, \uline{jumpable}, lovee, lumber-room, \textbf{merry-Andrew-like}, miliary fever, moonery, \textbf{mooning} (n.), namby-pamby, \textbf{puddingless}, sentimental, tittup
\end{quote}

Most of the words (14) are again nouns, followed by adjectives (5), verbs (1) and adverbs (1). Unlike the seventeenth century, the vast majority of the words have been formed within English through derivation (12), compounding (4) or conversion (2), and there are only two borrowings, \textit{dénouement} from French and \textit{hookah} from Arabic, and one verb of unclear etymology, \textit{tittup}. Looking at the top level of the HT classification and the main semantic class of each word, 9 of the neologism types relate to `the world' and 6 each to `society' and `the mind'. Within the classes, the words are again widely dispersed, but the most frequent categories are `the mind » emotion' and `society » leisure' at 3 types each. This paints a picture of letters written more for the purposes of keeping in touch and building friendships than for exchanging information, as in example \ref{ex:merryandrewlike}. This may be partly explained by the composition of the sample: in terms of the relationship between the sender and recipient, the largest category is letters between close friends (31\%), the proportion of which in the 17th-century sample is c.\ 21\%.

\ex. \label{ex:merryandrewlike}Your invocation has mounted me, \textbf{Merry Andrew-like}, upon stilts. -- I ape you, as monkeys ape men by walking upon two.\exref{(Ignatius Sancho to his friend, William Stevenson, 1777; SANCHO\_016)}

\subsection{Sociolinguistic Variation}

\begin{table}[t]
    \caption{Normalized frequency of neologism tokens per \num{10000} words in the 18th-century sample for the three main social axes of interest, sorted by frequency.}
    \label{tab:c18freqs}
    \centering
    \begin{tabular}{l@{\qquad}l@{}r}
    \toprule
    Category & Value & Frequency / \\
    & & \num{10000} words \\
    \midrule
    Gender & Male & \num{5} \\
    & Female & \num{4} \\
    \midrule
    Social Rank & Other non-gentry & \num{14} \\
    & Clergy & \num{7} \\
    & Nobility & \num{4} \\
    & Professionals  & \num{4} \\
    & Gentry & \num{3} \\
    & Royalty & \num{0} \\
    & Merchants & \num{0} \\
    \midrule
    Relationship & Close friends & \num{7} \\
    & Nuclear family & \num{5}\\
    & Other family & \num{3} \\
    & Other acquaintances & \num{1} \\
    \bottomrule
    \end{tabular}
\end{table}

In contrast to the 17th century, the \textbf{social rank} with the highest relative frequency of neologisms in the sample (14 tokens per \num{10000} words) is the lowest rank of all, other non-gentry; see Table \ref{tab:c18freqs}. Two out of the five individuals representing this rank use new vocabulary: Henry Barnes (\textit{foundling-house}) and Ignatius Sancho (\textit{blacky}, \textit{lovee}, \textit{merry-Andrew-like}, \textit{namby-pamby}). A husband to one of the nurses at the Foundling Hospital in London, Barnes addressed his letter to the matron of the hospital as in \ref{ex:foundlinghouse}. The hospital was founded in 1739 and seems to have been the first of its kind in England~\cite{Clark1994}, which explains the novelty of words referring to it.

\ex. \label{ex:foundlinghouse}For the meatrern of the \textbf{fondlen house}\exref{(Henry Barnes to Elizabeth Leicester, matron of the hospital, 1762;\\\mbox{}\hfill FOUNDLI\_126)}

Sancho, on the other hand, is quite a playful language user; see \ref{ex:merryandrewlike} above. He was an orphaned slave who ended up in England and found a position in the Duke of Montagu's household, later establishing a grocery shop~\cite{SanchoCharlesIgnatius17291780author}. He published several newspaper essays and rubbed shoulders with the literati of the time, and it seems that his social aspirations also show up in his use of neologisms.

In terms of \textbf{gender}, men tend to use more neologisms than women, as was also the case in the 17th-century sample. However, the proportion of female writers who use at least one new word is somewhat higher (5 out of 20) than the proportion of male writers (5 out of 33). The women come from the social ranks of gentry (Sarah Lennox and Hester Lynch Thrale), professionals (Frances Burney, author, and Mary Rawson Hart Boddam, wife of an East India Company employee, later governor, in Bombay), and clergy (Mary Colton, a vicar's wife and \textit{inspectress} of the Foundling Hospital). Both of the loanwords in the dataset again come from women: \textit{dénouement} was used by Hester Lynch Thrale to her literary friend, Frances Burney, discussing a plot, while Mary Rawson Hart Boddam described her new husband's smoking habits in Bombay as in \ref{ex:hookah}. For her part, Frances Burney wrote to her dramatist friend, Samuel Crisp, about herself as his \textit{anecdote-monger} and about her father's \textit{interference} in matters of matrimony, whereas Lady Sarah Bunbury née Lennox described a mutual relative as \textit{funny} to her friend, Lady Susan O'Brien.

\ex. \label{ex:hookah}He has been long at a subbordinate Factory and is a meer Moorman as to the language taste and customs, and will suck a Hubble Bubble, draw a Ailloon, smoak a \textbf{hooka} or \textbf{cream-cann} with you if you please; he has promised to wait on you with an account of us and to show you his different smoaking Machines being curious that way.\exref{(Mary Rawson Hart Boddam to her uncle, Thomas Pickering, 1760;\\\mbox{}\hfill DRAPER\_002)}

As for \textbf{register}, letters to close friends (as in \ref{ex:merryandrewlike} above) again show the most frequent use of neologisms, but this time they are followed by letters written to family members, and letters to other acquaintances like \ref{ex:foundlinghouse} are the least innovative register in this respect. In addition to Boddam above, clergyman Thomas Twining uses novel vocabulary when writing to his half-brother, Daniel, and Sir Roger Newdigate to his wife, Sophia. While Newdigate mostly writes with news from London like a friend having \textit{miliary fever}, Twining's letters are quite funny and aimed at entertaining the recipient, as in \ref{ex:puddingless}; the 16-year-old Daniel had tuberculosis and was staying at Bristol Hotwells for his health.

\ex. \label{ex:puddingless}Therefore, stick close to your Molière! for I shall set you a-translating again whenever I get you here; \& the vanquished shall be \textbf{pudding-less} for \textit{two} days, \& not have three puddings for it on the third!\exref{(Thomas Twining to his half-brother, Daniel Twining, 1764;\\\mbox{}\hfill TWINING\_005)}

Considering \textbf{age}, this time it seems that younger people under 40 use neologisms more frequently than those aged 40 and over (\num{5} vs.\ \num{4} tokens per \num{10000} words, respectively). If, however, we exclude the five new words by Thomas Twining, who has been shown to be an outlier in his prolific neologizing in the corpus as a whole~\cite{Saily2018b}, the numbers shift in favour of older people, as the younger group is left with only \num{3} tokens per \num{10000} words. In our sample, the oldest person using new words is the above-mentioned Sir Roger Newdigate at 54, whereas the youngest is Mary Rawson Hart Boddam at fifteen or sixteen. Another gentleman of middling years is Horace Walpole, politician and later Earl of Orford, who wrote at 51 to his friend, poet Thomas Gray, about a literary novelty \ref{ex:sentimental}.

\ex. \label{ex:sentimental}I think you will like Sterne's \textbf{sentimental} travels, which tho often tiresome, are exceedingly goodnatured \& picturesque.\exref{(Horace Walpole to his friend, Thomas Gray, 1768; GRAY\_065)}

As the 18th-century section of the corpus was not designed to be regionally representative, we shall omit \textbf{regional variation}, noting only that the young Mary Boddam serves as a good example of vocabulary acquired abroad; see \ref{ex:hookah} above.

\section{Discussion}

What do the samples tell us about our three social axes of interest? In terms of \textbf{gender}, male writers seem to lead the way, although the difference is less clear-cut in the eighteenth century. While there is very little previous sociolinguistic research on this, we may note a study by Keune~\cite{Keune2012}, who found that in Present-Day Dutch, the greatest lexical productivity was exhibited by highly educated older men, which also matches our results on age (unfortunately, we do not have enough data on the education of our informants to be able to use it as a category in the analysis). Furthermore, Säily and her collaborators have found that where there is variation in the productivity of certain nominal suffixes from Early Modern to Present-Day English, it is typically men who use them more productively than women~\cite{Saily2014,SailySuomela2017}. Säily links this to men's more nominal style, which has been observed in both historical and Present-Day English.

As for \textbf{social rank}, the results vary wildly by time period. This could be due to the small size of the samples, but it could also reflect the social history of the two periods: in the seventeenth century, access to education, specialized registers, and new things and ideas in general was typically only available to the upper and middling ranks, whereas the eighteenth century saw some improvements in this respect. Be that as it may, what we can say is that clearly it is not only men and the higher ranks who use new vocabulary, which implies that it is important to conduct more studies like ours which do not focus on published texts alone but which aim to represent a wider section of the populace. In a previous study, too, we found neologisms used by the lower ranks that were underdocumented in the OED, including \textit{Norfolker} `a Norfolk sheep or cow', a term that could merit inclusion in the OED, and an antedating to \textit{winterer} `an animal kept over the winter'~\cite{Saily2018b}. We have already submitted some of our antedatings to the OED, and entries are constantly being improved through ongoing work on the third edition of the dictionary.

\textbf{Register}, or the relationship between the writer and the recipient of the letter, has provided us with the most consistent results: it seems that letters between close friends are the most conducive to the use of new vocabulary. This is in accordance with Wolfson's ``bulge theory''~\cite{Wolfson1990}, which posits that a less stable relationship, such as that between status-equal friends, is subject to more negotiation and differs from both intimate family relationships and socially more distant relationships in terms of language use. Friends may wish to impress each other and put more effort into generating in-group solidarity, which could trigger more creative language use; cf. \cite{Saily2018a}.

What can we say about differences in neologism use between the two \textbf{time periods} represented by our samples? In 1640--1660, we are able to observe the ``Civil-War effect'' discussed by Raumolin-Brunberg~\cite{Raumolin-Brunberg1998} and Lijffijt et al.~\cite{Lijffijt2012}: the frequency of neologisms is much greater in this period than in the comparatively more tranquil 1760--1780, when wars were mostly fought further afield, and many of the 17th-century words are either directly or indirectly related to the war. We would argue that this phenomenon is not limited to the English Civil War but can be seen as part of what Dixon~\cite{Dixon1997} calls ``punctuated equilibria'', which is the notion that language history is characterized by periods of relative stability punctuated by external events that cause sudden changes in the linguistic situation and hence accelerate the rate of linguistic change. Nevalainen et al.~\cite{Nevalainen2020} have shown that in Middle English, such punctuating events included the Norman Conquest and the Black Death; at the lexical level, the legacy of the conquest shows up in the many French loanwords in English.

Comparing the most frequent semantic classes of the time periods, while the samples are small and admittedly biased, the change from `society » authority' to `society » leisure' and `the mind » emotion' may be indicative of a wider change in the styles and purposes of letter-writing. As noted by Somervell~\cite{Somervell2011}, in the eighteenth century ``letter-writing conventions became less formal, with their subject-matter including private as well as public matters, and letters were becoming an artistic, moral and intellectual literary form''. According to Säily~\cite[p.~214]{Saily2018a}, ``It is possible that letter-writing, at least for the upper classes in our corpus [i.e., the 18th-century section of the CEEC], was more about maintaining and building social relationships and identities than about conveying information.'' While the need to name things has always been one of the functions of neologism use (e.g. \textit{tea}, \textit{hookah}, \textit{foundling-house}), eighteenth-century letter-writers in particular seem to delight in using inventive terms to describe people and their actions (e.g. \textit{anecdote-monger}, \textit{lovee}, \textit{namby-pamby}, \textit{moonery}) or to discuss mutual interests like literature (e.g. \textit{dénouement}, \textit{sentimental}).

\section{Conclusions}

This paper has studied the users of new vocabulary and the uses to which they put it in two samples of English letters from the seventeenth and eighteenth centuries. We have found that while male letter-writers tend to use neologisms more frequently than women, the eighteenth century seems to have provided more opportunities for women and the lower ranks to participate in neologism use as well. In both samples, neologisms most frequently occur in letters written between close friends, which could be due to this less stable relationship triggering more creative language use. In the seventeenth-century sample, we have observed the influence of the Civil War, while the eighteenth-century sample appears to reflect the changing functions of letter-writing, as correspondence is increasingly being used as a tool for building and maintaining social relationships in addition to exchanging information.

In this study, we have only analysed words that could be mapped to the OED, thus focusing on lexical items that were at some point well-established enough in the language to warrant their inclusion in the dictionary. There are, however, a number of potential neologisms that are not listed in the OED at all; while some of these may be nonce-words, others could be more established words that have simply been missed by the OED editors, especially when it comes to the second edition, which was compiled in the nineteenth and early twentieth centuries. It is a well-known fact that the eighteenth century has been less well covered by the OED than the seventeenth century~\cite{Brewer2007}, and so it is perhaps no surprise that we find more of these neologism candidates there. Out of the words not mapped to the OED, there are 13 viable candidates in the eighteenth-century sample and only 4 in the seventeenth century. (It should be noted that even if these words were included in our analysis, the frequency of neologism use would still be higher in the seventeenth century, supporting our hypothesis of the Civil-War effect.) While many of them are Anglo-Indian words used by Mary Boddam and her sister, Eliza Draper, others are more general formations, such as \textit{fellow-labourer}, \textit{Pelhamized} and \textit{soul-cheering}. Whether these are mere nonce-formations could be assessed by tracking them in large databases of contemporary published texts, like Eighteenth Century Collections Online~\cite{ecco}.

As our quantitative approach has been based on comparing word forms (or strings separated by spaces) with those occurring earlier in the corpus as well as with the OED, we are missing many instances of homonyms, conversion, and multi-word units. According to the OED, however, the most frequent etymology types are borrowing and derivation~\cite[p. 351]{Nevalainen1999}, and we would expect these to be fairly well covered by our approach, so we may in fact have captured the majority of the neologisms in the samples. In future work, we could nevertheless attempt to improve our coverage of the other means of word-formation: as an example, it would be relatively easy to automatically compare bigrams in addition to unigrams to capture more of the compounds.

Despite the improvements in the normalization step \cite{hamalainen-etal-2019-revisiting}, normalization of the entire CEEC is still a problem that is far from solved. While using synthetic data improves low-resourced sequence-to-sequence models including character-level models \cite{hamalainen2019finding,hamalainen-wiechetek-2020-morphological}, our experiments with back-translation on the training data available to us have not yielded better accuracies. This is due to the fact that our training data comes from a different distribution and presents variation that is very different from the variation in the CEEC. The training data comes mainly from authoritative sources, such as historical forms recorded in the OED or the MED. This means that if any synthetic data is used, it should be representative of the historical variation in the CEEC itself. This, however, is not an easy task and it is something to be solved in the future.

In the course of our work, we have learned to ask more focused questions rather than attempting to analyse the corpus as a whole, as the amount of manual labour required to identify and classify neologisms is considerable. While neural machine-learning approaches may help with the normalization task necessary for identifying neologisms, it seems that to fully realize their potential, we would need much more data to be able to train better models. It is to be hoped that more and more letters and other egodocuments will be digitized in the future, as their contents and social representativeness are of interest to a great number of scholars in the digital humanities.

\section*{Acknowledgements}

We would like to thank the anonymous reviewers for helpful feedback. This work was supported in part by the Academy of Finland, Grants 293009 and 323390.

%
%
%
\bibliographystyle{splncs04}
\bibliography{mybibliography}

\end{document}